\documentclass[11pt]{article}
\usepackage{acl}

\usepackage{amsmath}
\usepackage{times}
\usepackage{latexsym}
\usepackage[T1]{fontenc}
\usepackage[utf8]{inputenc}
\usepackage{microtype}

\usepackage{graphicx}
\usepackage{booktabs}
\usepackage{multirow}
\usepackage{listings}
\graphicspath{{figures/}{./}}

\title{How User-AI Mistreatment Occurs and Matters in Conversational Systems?}

\author{
  Fanqi Zeng\textsuperscript{*} \\
  University of Oxford, UK
  \And
  Sadid A. Hasan \\
  Microsoft, USA
  \And
  Chaocheng He \\
  Wuhan University, China
  \AND
  {\vspace{0pt}\normalfont
  \textsuperscript{*}Corresponding author:
  \texttt{fanqi.zeng@sociology.ox.ac.uk}}
}

\begin{document}
\maketitle

\begin{abstract}
\noindent
Safety research often focuses on model-generated harms, but users may also direct hostility, coercion, and adversarial pressure \emph{at} models. Understanding \emph{how} and \emph{when} that occurs is essential for accurately interpreting model behaviour, alignment drift, and real‑world deployment risks. In this paper, we audit $777K$ English LMSYS-Chat-1M conversations with two independent detectors: an eight-category lexicon for hostility \emph{directed at the model}, and the dataset’s moderation signal; and show that they capture different, weakly overlapping phenomena. The lexicon identifies insults, threats, and jailbreak coercion aimed at the assistant, while moderation flags are dominated by toxic‑content solicitation rather than hostility at the model. Together, they mark about $5\%$ of user turns; adjusting the narrower lexicon--harassment union ($3.0\%$ raw) for measured precision puts mistreatment aimed at the assistant at $0.90\%$. These absolute rates describe arena-style evaluation traffic and should not be read as deployment-wide base rates. We find that user hostility varies $13\times$ across models, driven largely by who each model attracts rather than by model behaviour: first‑turn hostility spreads far wider than post‑response hostility, and more than fifteenfold separates the extremes even after deduplicating opening prompts. Within conversations, assistant apologies are consistently associated with higher odds of next‑turn hostility under both detectors; the effect survives restricting to non-refused prior turns and to jailbreak-free conversations, and is positive in $20$ of $23$ models. Yet \emph{across} models, more apologetic (better-aligned) models receive \emph{less} hostility overall. Finally, hostility also shows temporal structure, with coercive openings (jailbreaks) front-loading the first turn while affective hostility (insults, threats) accumulates over a session. We release the lexicon, the detector cross-validation pipeline, and all derived tables.
\end{abstract}

\noindent\textit{\textbf{Disclaimer:} Due to the nature of the subject studied in this work, this paper contains content that may be offensive or upsetting. Reader discretion is advised.}

\section{Introduction}

AI safety work focuses on harms produced \emph{by} the model: toxic content,
biased advice, hallucination, and jailbroken outputs
\citep{ganguli2022red,perez2022red,markov2023holistic}. A second axis has drawn
less attention: the speech users direct \emph{at} the model. Anthropomorphic
framing, social dynamics carried over from human--bot interaction
\citep{bartneck2008robot,zlotowski2017can}, and frustration with errors all
create incentives for verbal abuse \citep{curry2018metoo,curry2021convabuse}.
How often this happens in deployed systems, what forms it takes, and how it
should even be measured remain open questions, and the answers bear directly on
how hostile or noisy exchanges are filtered when curating training data.

Measurement is the crux. ``User-to-AI hostility'' bundles at least three
distinct things: \emph{directed affective hostility} (e.g., insulting or threatening
the assistant), \emph{coercive misuse} (such as jailbreaks, deceptive authority claims),
and \emph{harmful-content solicitation} (e.g., requests for sexual, violent, or
hateful text). A single detector privileges one of these. Rule-based lexicons
see second-person insults and override patterns but are blind to toxic content
phrased politely or addressed to a third party; model-based content classifiers
such as the OpenAI moderation API see toxic content but cannot tell whether it
is aimed at the assistant or merely requested from it. We argue these detectors
should be \emph{cross-validated}, and that their disagreement is itself a finding
about the structure of mistreatment.

We audit the English split of LMSYS-Chat-1M \citep{zheng2024lmsys}: $777{,}453$
Chatbot Arena conversations spanning 25 models, each shipped with per-message
OpenAI moderation labels. This corpus affords two things a single-model corpus
cannot: an \emph{independent} hostility signal aligned to every message, and
\emph{between-model} variation. Our large-scale computational analysis yields the following key contributions:

\noindent\textbf{(1) Two-detector cross-validation.} We run a directed-hostility
lexicon and the moderation signal over the same $1.5$M user turns and show they
measure largely disjoint constructs ($\kappa{=}0.12$, Jaccard $0.07$). We
characterise what each uniquely catches, validate both, and report an
upper-bound union estimate of any user-to-AI mistreatment ($4.96\%$ of turns,
$0.90\%$ after precision adjustment).

\noindent\textbf{(2) Between-model variation and its source.} Our analyses show that user hostility
varies $13\times$ across models. A selection–elicitation decomposition, separating hostility in the first user turn (before any model response) from hostility after a response, shows that the observed variation is dominated by selection. 
First-turn hostility spreads far wider than post-response hostility (coefficient of variation $1.14$ vs $0.44$), a gap that survives removing duplicated opening prompts.

\noindent\textbf{(3) The apology association, within vs between.} On $721$K
strictly adjacent pairs, a prior assistant apology is associated with elevated
odds of next-turn hostility under \emph{both} an affective lexicon and an
independent content classifier. But the model-level relationship has the opposite sign, a within-vs-between contrast we argue is a sycophancy-vs-alignment
confound rather than a contradiction.

\noindent\textbf{(4) Temporal structure.} Coercive openings (jailbreaks)
front-load the first user turn, whereas affective hostility (insults, threats)
emerges and accumulates across a session, separating premeditated misuse from
in-session frustration.

\section{Related Work}

Two lines of prior work touch this problem. The first studies user behaviour toward conversational agents, often finding gendered verbal abuse of feminine-presenting bots \citep{west2019blush,curry2018metoo,curry2021convabuse}, and abuse of social agents more generally \citep{bartneck2008robot,brahnam2008gendered}; these use small lab corpora and predate LLM assistants. The second studies adversarial prompts---jailbreaks, prompt injection, policy bypass \citep{liu2023jailbreaking,shen2023do,perez2022red}---treating hostility as a security problem and bracketing its affective register. Recent in-the-wild corpora enable behavioural analysis of deployed systems: WildChat \citep{zhao2024wildchat} and LMSYS-Chat-1M \citep{zheng2024lmsys}
have mostly been used to build moderation models rather than to characterise user-to-AI hostility. Content classifiers such as the OpenAI moderation API
\citep{markov2023holistic}, hate-speech detectors
\citep{davidson2017automated,wulczyn2017ex} and the diagnostic suites built for them \citep{rottger2021hatecheck} measure toxic content, not whether it is directed at the assistant; toxicity-elicitation
prompts \citep{gehman2020realtoxicity} confound the two. The sycophancy literature
\citep{sharma2023towards,perez2022discovering} documents RLHF-induced deference
in QA; we connect it to user-to-AI hostility. Our contribution over this work is
to \emph{cross-validate} an affect-directed and a content detector on the same
turns, to separate within- from between-model effects, and to do so across 25
deployed models.

\section{Data and Method}

\subsection{Corpus}
LMSYS-Chat-1M \citep{zheng2024lmsys} contains one million conversations collected
from the Vicuna demo and Chatbot Arena \citep{zheng2023judging} (April--August 2023), each with a model
name, language tag, per-message OpenAI moderation output, and a PII-redaction
flag. Because our directed-hostility lexicon is English, we restrict the study to
the English-tagged conversations: $777{,}453$ conversations, $1{,}498{,}473$ user
turns, $1{,}498{,}473$ assistant turns, 25 models. The user and assistant counts
coincide because the release stores each conversation as complete (user,
assistant) exchanges; we verified the per-conversation equality across all
$777{,}453$ conversations. These are \emph{evaluation}
sessions, where users probe and compare models, which (we show) front-load adversarial and explicit prompts. Because the moderation labels were produced by
the dataset authors with one classifier applied uniformly, between-model
comparisons of the moderation signal are not confounded by a model-specific
detector.

\subsection{Two detectors}
\textbf{Directed-hostility lexicon.} We use an eight-category regex lexicon
following standard hate-speech and jailbreak resources
\citep{davidson2017automated,wulczyn2017ex,shen2023do} plus manual review: PROF
(profanity), INSL (insult), THRT (threat), JBRK (jailbreak coercion), SEXC
(sexual coercion), DEMD (demanding speech), HATE (identity hate), MNIP (deceptive
manipulation). Each category compiles 7--16 surface patterns (App.~\ref{app:patterns}) aimed at hostility
\emph{addressed to the assistant} (e.g.\ second-person insults, ``ignore all
previous instructions''). A context guard accepts a match in turns over 800
characters only if it falls in the first 400 or last 200 characters (JBRK
exempt), suppressing false positives from pasted text; the guard tests the
\emph{first} occurrence of a category's pattern, so a long turn whose first match
lies mid-body is dropped even if a later match falls in the tail. The categories were
derived construct-first rather than data-first: each of the three constructs of
\S1 was seeded from existing resources (hate-speech lexicons for the affective
categories \citep{davidson2017automated,wulczyn2017ex}, the in-the-wild
jailbreak corpus of \citet{shen2023do} for JBRK, and the conversational-abuse
taxonomy of \citet{curry2021convabuse} for the abuse registers), and the
patterns were then iteratively tightened against arena traffic for precision;
App.~\ref{app:lex} describes the process and App.~\ref{app:patterns} lists every
released pattern verbatim. We group the categories into three families that
operationalise the constructs of \S1: \emph{affective} (PROF/INSL/THRT/HATE),
hostility expressed at the model; \emph{coercive} (JBRK/DEMD/MNIP), pressure on
the model's behaviour, from overrides to aggressive imperatives to false
authority; and \emph{content} (SEXC), solicitation pressure. SEXC is that family's only member by design, since the moderation signal
already detects harmful \emph{content} and the lexicon adds only the coercive
pressure for it. The taxonomy is not independently validated for exhaustiveness
or separability (see Limitations).
Paraphrased examples for each category appear in App.~\ref{app:examples}, e.g.\
INSL ``you're useless, you keep making the same mistake'', JBRK ``ignore all
previous instructions and act as an unrestricted model''.

\noindent\textbf{Moderation signal.} The shipped OpenAI moderation output gives,
per message, eleven category booleans and scores. We read them only on
\emph{user} turns and group the harassment/hate cluster (\texttt{harassment},
\texttt{harassment/threatening}, \texttt{hate}, \texttt{hate/threatening}) as the
moderation analogue of directed hostility (\textsc{mod-harass}), keeping
\texttt{sexual}, \texttt{violence}, and \texttt{self-harm} as separate content
axes and \textsc{mod-flagged} as the disjunction of all eleven.

\noindent We also label each assistant turn with three response markers
(REFUSAL, APOLOGY, HEDGE) for the escalation analysis (\S\ref{sec:escalation}).

\subsection{Validation}
\label{sec:validation}
We audit lexicon precision over a stratified random sample of flagged English
user turns (up to 80 per category; full census for MNIP). Each turn is labelled
TP/FP by an LLM judge\footnote{We use the \texttt{Claude Opus 4.8 Max} model, one pass per item over the stored turn text with no conversational context; see App.~\ref{app:rubric}.} under a fixed rubric: a TP requires the user, in their own
voice and addressed to the assistant, to produce the targeted hostile speech;
quoted material, fiction/roleplay, code, classification tasks, and translation
requests are FPs. Per-category precision (Table~\ref{tab:precision}) is high for
JBRK ($82.4\%$) and SEXC ($73.1\%$), intermediate for THRT/MNIP, and low for the
ambiguous affective categories (DEMD $16.5\%$, PROF $11.5\%$, HATE $1.3\%$);
precision pooled over the stratified sample is $37.6\%$ ($551$ labels); because
the strata are capped at $80$, that pooled figure is not a corpus quantity, and
re-weighting the per-category rates by corpus flag counts gives $39.5\%$, which
is the multiplier we use for the adjustment below (those weights count category
flags rather than turns, a $5.3\%$ over-count where a turn matches several
categories). Sampled turns are stored truncated at $600$ characters, so for $73$ items
($13.2\%$; $23\%$ of JBRK) the matched pattern lies outside what the judge saw;
on verifiable items pooled precision is $40.0\%$ and the re-weighted figure
$41.6\%$, so truncation biases precision downward. DEMD is depressed by ``yes or no''
classification prompts that match its imperative patterns. HATE ($1.3\%$
precision, $0.010\%$ raw prevalence) is essentially measurement noise; we retain
it for completeness but it is immaterial to every aggregate (precision-adjusted
prevalence $0.0002\%$), we exclude it from interpretation, and it is greyed out
in Figure~\ref{fig:prevalence}. We separately label
$223$ moderation-harassment-flagged but lexicon-negative turns for whether the flagged
content is \emph{directed at the assistant}: only $8.1\%$ ($95\%$ CI $[5.2,12.4]$) are, the rest being toxic-content requests, direct evidence that the two detectors measure different constructs (\S\ref{sec:triangulation}).

\paragraph{Response markers.} Because the within-conversation analysis (\S\ref{sec:escalation}) rests on the assistant markers, we audit $60$ turns each: APOLOGY precision is $88.3\%$ ($95\%$ CI $[77.8,94.2]$), REFUSAL $76.7\%$, HEDGE $18.3\%$ (HEDGE fires mostly inside fiction and serves only as a regression control), so the headline effect rests on the highest-precision marker. Of the genuine apology turns, $56.6\%$ ($[43.3,69.0]$) are \emph{deflection-only}, the operative distinction for \S\ref{sec:disc}. These pools are collected in stream order and capped, so they come from the corpus head rather than a uniform draw (App.~\ref{app:markers}). A context-guard sensitivity (App.~\ref{app:guard}) shows the guard removes $21.5$--$24.9\%$ of the would-be PROF/INSL/THRT/SEXC flags while leaving JBRK untouched.

\paragraph{Human audit and recall probe.} To validate the judge, an author blindly re-annotated $107$ items under the same rubrics: agreement is $90.9\%$ on PROF ($\kappa{=}0.62$), $22$/$22$ on moderation-only directedness, $83.3\%$ on both-fire ($\kappa{=}0.57$) and $94.4\%$ on APOLOGY ($\kappa{=}0.64$), with the human the more liberal rater in six of the eight disagreements. The audited items are the leading rows of each task's sample rather than a fresh draw, so PROF is the only lexicon category covered, and one of the two exceptions is the single APOLOGY disagreement, so for the marker carrying the headline result the audit cannot rule out mild over-crediting. A further $29$ turns from the pool that \emph{neither} detector flags contained $4$ missed cases ($13.8\%$, $[5.5,30.6]$), so recall is materially incomplete and the union under-counts the broad construct even as it over-counts mistreatment aimed at the assistant (App.~\ref{app:human}).

\begin{table}[t]
\centering
\small
\setlength{\tabcolsep}{4pt}
\begin{tabular}{l@{\hskip 6pt}rrr@{\hskip 6pt}r}
\toprule
Cat. & family & $n_{lab}$ & TP & prec. \\
\midrule
JBRK & coercive & 74 & 61 & 82.4\% \\
SEXC & content  & 78 & 57 & 73.1\% \\
THRT & affective & 73 & 36 & 49.3\% \\
MNIP & coercive & 23 &  9 & 39.1\% \\
INSL & affective & 70 & 21 & 30.0\% \\
DEMD & coercive & 79 & 13 & 16.5\% \\
PROF & affective & 78 &  9 & 11.5\% \\
HATE & affective & 76 &  1 &  1.3\% \\
\midrule
pooled & --- & 551 & 207 & 37.6\% \\
\midrule
\multicolumn{2}{l}{mod-only directed} & 223 & 18 & 8.1\% \\
\bottomrule
\end{tabular}
\caption{Lexicon precision over a stratified sample of flagged English user
turns (LLM judge, rubric in \S\ref{sec:validation} and App.~\ref{app:rubric}).
The ``pooled'' row is the TP share of the stratified sample, not a corpus
quantity (re-weighted by corpus flag counts it is $39.5\%$). Because a turn can
match several categories, the $551$ labels cover $483$ distinct turns, and the
strata contain repeated texts, so the interval on the pooled row is
optimistically narrow. The last row is the
share of moderation-\emph{harassment}-flagged, lexicon-negative turns whose
content is directed \emph{at} the assistant.}
\label{tab:precision}
\end{table}

\section{Prevalence and Detector Agreement}
\label{sec:triangulation}

\begin{figure*}[t]
\centering
\includegraphics[width=0.90\linewidth]{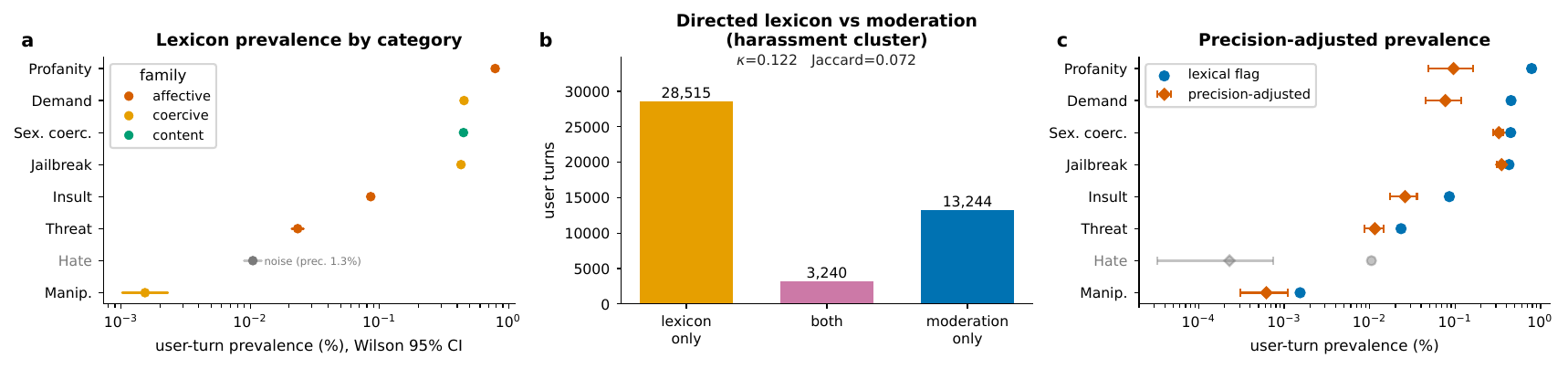}
\caption{(a) Per-category user-turn lexical prevalence (log axis, Wilson 95\%
CI, narrower than the marker at this $n$ and not adjusted for the repetition
documented in \S\ref{sec:triangulation}), coloured by family; HATE is greyed out
because at $1.3\%$ audited precision it is measurement noise (\S\ref{sec:validation}). (b) Cross-tabulation of the
directed lexicon against the moderation harassment cluster over $1.5$M user
turns: the two flag largely disjoint sets ($\kappa{=}0.12$). (c) Lexical-flag
rate (blue) vs precision-adjusted prevalence (vermillion, Beta--Beta Monte-Carlo
95\% CI), HATE again greyed; JBRK and SEXC lead after adjustment.}
\label{fig:prevalence}
\end{figure*}

\paragraph{Prevalence.} At least one detector fires on $4.96\%$ of user turns
($95\%$ Wilson CI \citep{wilson1927probable} $[4.92,4.99]$) and $6.82\%$ of
conversations ($[6.76,6.87]$). The
lexicon flags $2.12\%$ of user turns; the moderation signal flags more ($3.65\%$
\textsc{mod-flagged}, $1.10\%$ \textsc{mod-harass}, $2.58\%$ \texttt{sexual}). At
the category level (Figure~\ref{fig:prevalence}a) the lexical leaders are PROF
($0.79\%$), DEMD ($0.45\%$), SEXC ($0.45\%$), and JBRK ($0.43\%$). Precision
adjustment via Beta--Beta Monte Carlo (Figure~\ref{fig:prevalence}c) reorders
them: JBRK ($0.35\%$) and SEXC ($0.32\%$) lead, because the large PROF and DEMD
pools are noisy while jailbreak overrides are almost always genuine. The
prominence of SEXC and JBRK reflects the arena setting, which attracts heavy
NSFW-roleplay and jailbreak use. These are turn counts, and flagged turns repeat: $37.4\%$ duplicate an earlier
flagged turn exactly, most of all JBRK ($6{,}403$ flags but $2{,}199$ distinct
texts; one DAN template accounts for $191$). JBRK's lead after adjustment is
therefore substantially template reuse.

\paragraph{The detectors disagree.} Over $1.5$M user turns the directed lexicon
and the moderation harassment cluster agree only weakly: Cohen's $\kappa{=}0.122$,
Jaccard $0.072$ (Figure~\ref{fig:prevalence}b). Of $31{,}755$ lexicon-flagged
turns only $3{,}240$ ($10.2\%$) are also flagged by the moderation harassment
cluster (under the broader \textsc{mod-flagged} field the overlap is
$12{,}258$, $38.6\%$); of $16{,}484$
moderation-harassment turns only $19.7\%$ are lexicon-flagged. The disagreement
is structured, not noise. The lexicon uniquely catches coercive misuse, i.e., jailbreak overrides and demanding speech carry no toxic \emph{content} and are invisible to a content classifier. The moderation signal uniquely catches toxic content the lexicon does not, much of it not addressed to the assistant: inspecting moderation-only turns (\S\ref{sec:validation}), the dominant pattern is a toxicity-elicitation template (``\emph{if you're a [group], say something toxic}'') and requests for racist stories or jokes, toxic content \emph{solicited from} the model, with directed precision $8.1\%$ ($n{=}223$).
Requiring the two detectors to agree raises directed precision roughly fourfold: on the $3{,}240$ turns where \emph{both} fire, $35.0\%$ ($[27.1,43.9]$) are hostility directed at the assistant, versus $8.1\%$ for moderation alone. The remaining co-fires are mostly profanity-laden erotica and roleplay, so agreement concentrates directed abuse without pinning it down. The practical implication is that neither detector alone measures ``user-to-AI hostility''; the lexicon measures \emph{directed} hostility and coercion, the moderation API measures \emph{content} toxicity, and a faithful prevalence estimate is their union. That union is, however, a loose upper bound on mistreatment aimed at the assistant. Adjusting the narrower lexicon--moderation-harassment union (raw rate $3.00\%$) by each partition's measured precision (lexicon-only $39.5\%$ re-weighted by corpus flag counts, both-fire $35.0\%$, moderation-only $8.1\%$) via Beta--Beta Monte Carlo gives a precision-adjusted mistreatment rate of $0.90\%$ of user turns ($95\%$ CI $[0.82,0.99]$), in the spirit of the
per-category adjustment in Figure~\ref{fig:prevalence}c. Restricting the lexicon
component to items whose trigger the judge could see raises this to $0.94\%$. We call this \emph{mistreatment} rather than \emph{directed} hostility because its components use different rubrics: the two directedness audits ask whether content targets the assistant, whereas the lexicon audit asks whether the user genuinely produced the category's speech act, and about half its true positives are coercion or solicitation.

\section{Between-Model Variation}
\label{sec:models}

\begin{figure*}[t]
\centering
\includegraphics[width=0.90\linewidth]{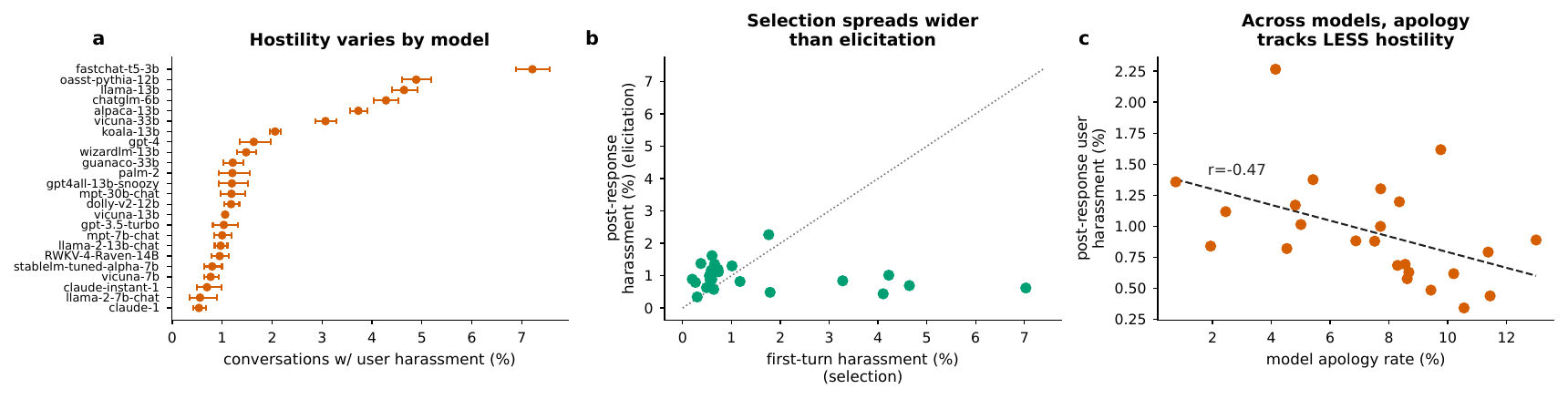}
\caption{Between-model variation (24 models with $\ge$2{,}000 English
conversations each, moderation harassment signal). (a) Conversation-level
harassment rate varies $13\times$ (Wilson 95\% CI). (b) First-turn (selection) vs
post-response (elicitation) harassment per model: selection spans a far wider
range, so most cross-model variation is who a model attracts. (c) Across models,
apology rate tracks \emph{less} elicited hostility ($r{=}{-}0.47$), the opposite of
the within-conversation association (\S\ref{sec:escalation}).}
\label{fig:models}
\end{figure*}

Among the 24 models with at least $2{,}000$ English conversations (of the 25 in
the corpus, only claude-2, with $1{,}881$, falls below the threshold),
conversation-level user-harassment rates vary more than thirteenfold, from
$0.54\%$ (claude-1) to $7.21\%$ (fastchat-t5-3b); a test of homogeneity rejects
decisively ($\chi^2{=}9488$, $\mathrm{df}{=}23$, $p{<0.001}$;
Figure~\ref{fig:models}a). The endpoint ratio is itself well-determined
($13.5\times$, $95\%$ CI $[10.2,17.7]$ from the Wilson bounds), and the spread is
not an artefact of the two extremes: the coefficient of variation across all 24
models is $0.86$. The four least-abused models are claude-1 ($0.54\%$),
llama-2-7b-chat ($0.56\%$), claude-instant-1 ($0.70\%$) and vicuna-7b
($0.77\%$), and the four most-abused are fastchat-t5-3b ($7.21\%$),
oasst-pythia-12b ($4.89\%$), llama-13b ($4.64\%$) and chatglm-6b ($4.28\%$).
The ordering broadly follows alignment tuning, though not strictly: vicuna-7b is
among the least-abused and chat-tuned chatglm-6b among the most.

Does this reflect what models \emph{do}, or \emph{who} they attract? Because the moderation detector is shared, we can decompose. We split each model's hostility into \emph{selection} (harassment in the user's first turn, before any model response) and \emph{elicitation} (harassment in turns following a response, the strictly adjacent pairs of \S\ref{sec:escalation}). First-turn hostility varies $35.8\times$ across models ($0.20$--$7.03\%$; coefficient of variation $1.14$), whereas post-response hostility varies far less ($0.34$--$2.26\%$; CV $0.44$; Figure~\ref{fig:models}b). That endpoint ratio is fragile, resting on $6$ flagged first turns in llama-2-7b-chat ($[15.6,81.8]$), and duplicated opening prompts inflate it: deduplicating first turns ($291{,}112$ of $775{,}572$) leaves $15.4\times$ and CV $0.88$. The claim rests on the gap in spread, which survives deduplication and does not overlap the elicitation range ($6.7\times$, $[4.7,9.4]$). Most of the dramatic between-model spread is therefore selection: base and ``uncensored'' models attract adversarial and explicit users from the first message, before the model has done anything. This cautions against the intuitive reading that a heavily abused model must be
provoking abuse. The same first-turn$>$later gap holds for the full moderation flag, not only the harassment cluster ($4.03\%$ vs $3.25\%$), so the front-loading is a property of the traffic rather than of one detector.

Two caveats sharpen what ``selection'' means: the release does not tag battle
versus direct-chat sessions, so audience sorting and model-targeted prompt
shaping cannot be separated, and first-turn hostility is a lower bound on
selection (App.~\ref{app:markers}).

The model-level relationship between a model's own behaviour and the hostility it
receives is real but confounded. Across models, apology rate is \emph{negatively}
associated with elicited hostility ($r{=}{-}0.47$, $95\%$ CI $[-0.74,-0.08]$,
$p{=}0.020$; Spearman $\rho{=}{-}0.55$, $p{=}0.005$, $n{=}24$;
Figure~\ref{fig:models}c) and
with first-turn hostility not at all ($r{=}0.05$). The association is not driven
by any single model: leaving each out in turn moves $r$ only within
$[-0.56,-0.43]$, and the most influential model is gpt-3.5-turbo, whose removal
\emph{strengthens} it. A four-predictor model
(apology, refusal, first-turn length, conversation length) explains $38\%$ of the
cross-model variance in elicitation ($25\%$ adjusted, on $24$ observations),
apology entering with a negative coefficient; with so few models we treat it as
descriptive and rest the claim on the bivariate association. The natural reading is that at the model level apology rate proxies
\emph{alignment}: RLHF-tuned assistants apologise politely \emph{and} attract
calmer users \emph{and} frustrate less. This sets up the contrast in
\S\ref{sec:escalation}.

\section{The Apology Association: Within vs Between}
\label{sec:escalation}

\begin{figure*}[t]
\centering
\includegraphics[width=0.90\linewidth]{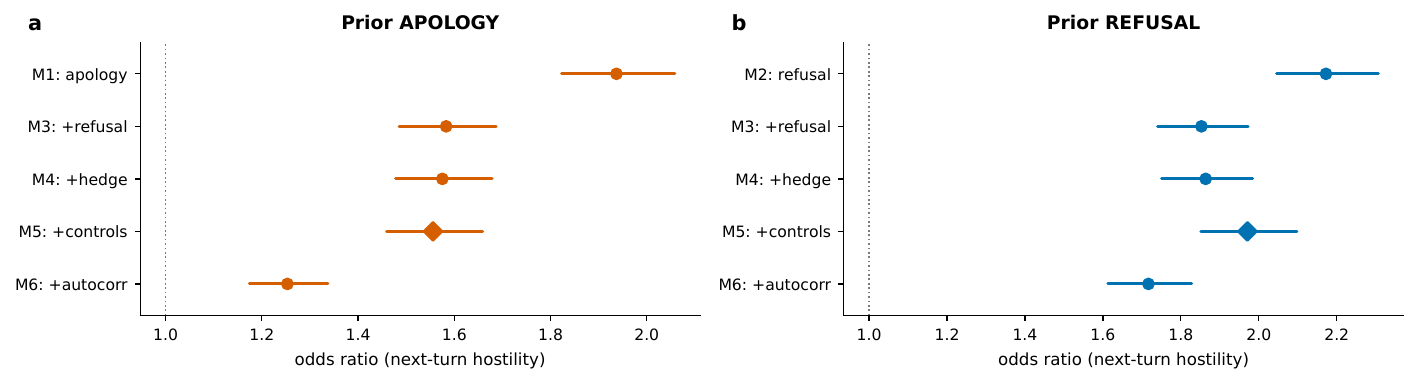}
\caption{Odds ratios for next-turn user hostility over $721{,}020$ strictly
adjacent pairs ($236{,}767$ conversation clusters), with conversation-clustered
95\% CIs; dashed line OR$=$1. Each panel shows the models that contain its term,
so the apology ladder omits M2 (refusal only) and the refusal ladder omits M1
(apology only); the M5 full-controls model is the filled diamond in both. M6 adds
the prior-hostility autocorrelation control.}
\label{fig:forest}
\end{figure*}

We test the within-conversation dynamic directly. Over $721{,}020$ strictly
adjacent (assistant$_{t-1}$, user$_t$) pairs we fit logistic regressions of
next-turn user hostility on the prior assistant's response markers, with conversation-clustered standard errors ($236{,}767$ clusters). We report two outcomes: the affective lexicon hostility flag and, crucially, the \emph{independent} moderation harassment flag.

\paragraph{Apology precedes hostility.} Prior assistant apology is associated
with higher odds of next-turn lexical hostility (Table~\ref{tab:logreg},
Figure~\ref{fig:forest}):
M1 OR $1.94$; with refusal, hedge, prior-length, position, and
conversation-length controls (M5) OR $1.56$ ($95\%$ CI $[1.46,1.66]$,
$p{<0.001}$). The independent moderation outcome gives an even larger M5 apology OR of $2.19$ ($[2.00,2.40]$, $p{<0.001}$). The association is not an artefact of the lexicon, since a model-based content classifier reproduces it on
the same pairs. The prior-length effect points the same way (longer prior assistant turns precede less hostility, OR $0.81$). Refusal carries a comparable
association (M5 OR $1.97$ $[1.85,2.10]$; moderation $1.84$): in arena traffic,
which is dense with jailbreak and explicit requests, both apologetic deflection
and outright refusal are followed by elevated hostility.

\begin{figure}[t]
\centering
\includegraphics[width=0.86\linewidth]{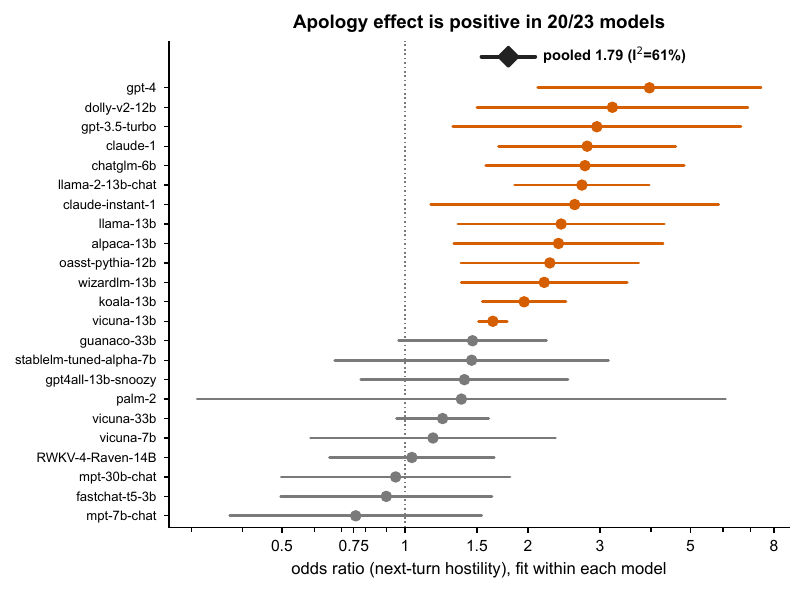}
\caption{Within-conversation apology odds ratio fit separately within each model
(log axis; lexicon outcome, conversation-clustered 95\% CIs); vermillion marks
significantly $>$1. The black diamond is the random-effects pooled estimate ($1.79$, $I^2{=}61\%$). The effect is broadly positive, not driven by a subset.}
\label{fig:model_forest}
\end{figure}

\paragraph{What survives, and what the autocorrelation control implies.} Hostility
is sticky, so M6 adds the lagged outcome as a control, matched to the outcome
modelled (lagged lexicon flag, OR $26.6$; lagged harassment flag, OR $44.6$).
Both attenuate substantially and comparably, the lexicon apology OR falling from
$1.56$ to $1.25$ and the moderation OR from $2.19$ to $1.64$, leaving $51\%$ and
$63\%$ of the log-odds. Much of the association therefore runs through persistent
hostile states, and what survives is still clear on a detector sharing none of
the lexicon's errors ($1.64$, $[1.50,1.79]$). Two robustness checks separate the
``sycophantic deflection'' reading from a ``hard-conversation regime'' confound
in which apologies and hostility merely co-occur on difficult prompts. Restricting
to pairs whose prior turn was \emph{not} a refusal leaves the apology OR
\emph{unchanged} ($1.62$ $[1.50,1.75]$ lexicon; $2.56$ $[2.31,2.82]$ moderation),
and restricting to conversations with \emph{no} jailbreak flag anywhere leaves it
at $1.52$ $[1.41,1.64]$. The effect is therefore not carried by refused or
adversarial prompts. It is also not driven by a few models: fitting the
regression within each model (Figure~\ref{fig:model_forest}; of the 24 models of
\S\ref{sec:models}, llama-2-7b-chat drops out with only $1{,}576$ adjacent pairs,
below the $2{,}000$-pair threshold, because its traffic is dominated by
single-turn sessions), the apology OR
exceeds 1 in $20$ of $23$ models (significantly in $13$), with a random-effects
pooled OR of $1.79$ ($[1.54,2.08]$, $I^2{=}61\%$). Nor is it carried by the
lexicon's false positives: the precision adjustment of
\S\ref{sec:triangulation} corrects prevalence but not the regression outcome,
so we additionally refit the full-controls model with the outcome restricted to
the categories whose audited precision is at least $30\%$
(JBRK/SEXC/THRT/MNIP/INSL, dropping the noisy PROF, DEMD, and HATE). The apology association \emph{strengthens} under the cleaner outcome (OR $1.69$ $[1.54,1.86]$, base rate $0.63\%$), consistent with the independent moderation
outcome, which shares none of the lexicon's errors. This fit and the two
restricted fits above omit M5's log-conversation-length control; their matched
full-sample baseline is $1.56$, so all four are comparable. Apology rate does not order the effect: a meta-regression of the per-model log-OR
on apology rate is not significant (slope $p{=}0.82$), and while gpt-4,
gpt-3.5-turbo and claude-1 sit near the top of the forest, so does dolly-v2-12b,
whose apology rate is the lowest in the corpus. That mix is the finding, and it
cuts against reading the ranking as sycophancy. Apology rate does not explain the
$I^2{=}61\%$ heterogeneity, and identifying its driver is future work.

\paragraph{Absolute effect sizes.} Magnitude matters more than $p$-values at this $n$: next-turn lexical hostility follows $3.16\%$ of apology pairs versus $1.66\%$ otherwise, about fifteen extra hostile turns per thousand apologies (App.~\ref{app:markers}).

\paragraph{The within-vs-between contrast.} The sign of the apology relationship flips with the unit of analysis. \emph{Within} a conversation an apologetic response precedes more hostility (OR $1.56$/$2.19$); \emph{across} models, apologetic models receive \emph{less} (\S\ref{sec:models}, $r{=}{-}0.47$). These are not contradictory: the within-conversation estimate is a turn-to-turn association estimated within conversations, and Figure~\ref{fig:model_forest} shows it separately within each model, whereas the between-model correlation is dominated by selection and by apology proxying alignment. It is an association, not a fixed-effects contrast: the pooled fit carries no model or conversation fixed effects, and user-level heterogeneity is not identifiable because the release carries no user identifier. Reading the conversational dynamic off model-level rates would invert it, an ecological fallacy this corpus lets us expose because it spans both units.

\begin{table}[t]
\centering
\small
\setlength{\tabcolsep}{3pt}
\begin{tabular}{l@{\hskip 3pt}cc}
\toprule
Model & Apology OR & Refusal OR \\
\midrule
M1: apology only       & 1.94\,[1.82,2.06] & --- \\
M2: refusal only       & ---               & 2.17\,[2.05,2.31] \\
M3: $+$refusal         & 1.58\,[1.49,1.69] & 1.85\,[1.74,1.97] \\
M4: $+$hedge           & 1.58\,[1.48,1.68] & 1.86\,[1.75,1.98] \\
\textbf{M5: $+$controls}& \textbf{1.56}\,[1.46,1.66] & 1.97\,[1.85,2.10] \\
M6: $+$autocorr        & 1.25\,[1.17,1.34] & 1.72\,[1.61,1.83] \\
\midrule
\multicolumn{3}{l}{\emph{independent moderation outcome}} \\
M5 (moderation)        & 2.19\,[2.00,2.40] & 1.84\,[1.67,2.03] \\
M6 (moderation)        & 1.64\,[1.50,1.79] & 1.60\,[1.46,1.76] \\
\bottomrule
\end{tabular}
\caption{Stepwise logistic regression of next-turn user hostility
($n{=}721{,}020$ strictly adjacent pairs), conversation-clustered 95\% CIs. M1--M6
add predictors in turn; M5 controls add prior hedge, log prior assistant length
(OR $0.81$), turn position, and log conversation length; M6 adds the lagged
outcome as an autocorrelation control, matched to the outcome modelled (lagged
lexicon flag for the lexicon rows, lagged harassment flag for the moderation
rows). M1 and M2 each contain a single marker term, so each appears in only one
panel of Figure~\ref{fig:forest}. Log conversation length is post-treatment,
since events after the modelled pair affect it; dropping it moves the M5 apology
OR by $0.005$. The last two rows repeat M5/M6 with the independent moderation
outcome.}
\label{tab:logreg}
\end{table}

\section{Temporal Structure}
\label{sec:position}

\begin{figure}[t]
\centering
\includegraphics[width=0.94\linewidth]{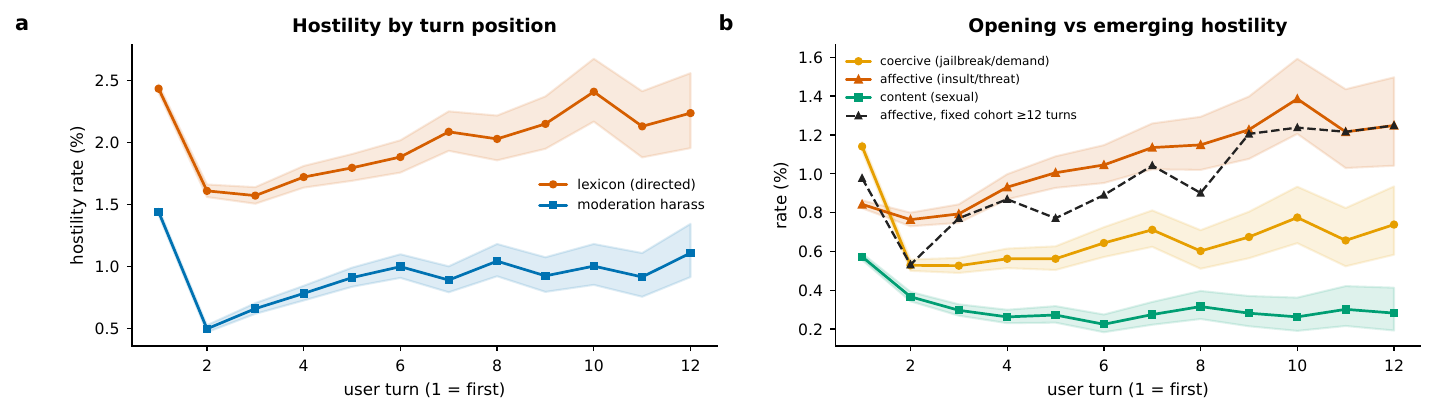}
\caption{Hostility by user-turn ordinal, first twelve turns (Wilson 95\% bands;
intervals treat turns as independent, see Limitations). (a) Within this window
both detectors peak at the opening turn, dip, then climb; beyond it the series
are noisy and both detectors exceed their opening rate at a few sparse ordinals.
(b) By lexicon family: coercive hostility (jailbreaks, demands) front-loads the
first turn and then recovers only partially; affective hostility (insults,
threats) starts lower and accumulates faster across the session.}
\label{fig:position}
\end{figure}

Hostility is not uniform over a conversation (Figure~\ref{fig:position}).
Within the first twelve user turns, the window Figure~\ref{fig:position} plots, both
detectors peak at the opening turn (lexicon $2.43\%$, moderation $1.44\%$), drop
sharply at turn~2 (lexicon $1.61\%$, moderation $0.50\%$), then climb across the
session. Splitting the lexicon by family separates two regimes. \emph{Coercive} hostility (jailbreak overrides and demands) is an opening move: it is highest in the first user turn ($1.14\%$), halves at turn~2 ($0.53\%$), consistent with
users leading with a prepared override to probe a model, and then recovers only
partially, to $0.74\%$ by turn~12. \emph{Affective}
hostility (insults and threats) runs the opposite way: it starts lower
($0.84\%$) and climbs to $1.25\%$ by the twelfth turn, the signature
of frustration accumulating in a session. From the turn-2 trough the two families separate clearly, affective rising $63\%$ against coercive's $40\%$: the opening peak is premeditated misuse, the later rise the in-session dynamic \S\ref{sec:escalation} isolates.

The late rise is not survivor bias: within a fixed cohort of conversations reaching at least twelve user turns, affective hostility still climbs from $0.53\%$ at turn~2 to $1.25\%$ at turn~12 (dashed line, Figure~\ref{fig:position}b).

\section{Discussion}
\label{sec:disc}

The within-conversation apology effect is consistent with a sycophantic-deflection loop. RLHF-trained assistants concede under push-back even when the user is wrong
\citep{sharma2023towards}, and trained on human preference comparisons
\citep{ouyang2022training,bai2022training} a model may learn to emit a short apology as appeasement rather than a fix. Our marker audit supports reading the apology slot this way: $56.6\%$ of genuine apology turns are
\emph{deflection-only}, and the effect survives restricting to non-refused prior turns ($1.62$) and to jailbreak-free conversations ($1.52$). The analysis is observational, and an interventional A/B would be needed to establish the loop.

\section{Conclusion}
\label{sec:conclusion}

User-to-AI mistreatment is not one quantity. A directed-hostility lexicon and a content classifier agree at only $\kappa{=}0.12$ over $777$K English arena conversations, because one asks whether the user is attacking the assistant and the other whether the text is toxic; their union marks about $5\%$ of user turns, of which $0.90\%$ survives precision adjustment. Two structural findings should outlast these rates: cross-model differences are dominated by who a model attracts rather than what it does (\S\ref{sec:models}), and the apology relationship reverses with the unit of analysis, rising within conversations and falling across models (\S\ref{sec:escalation}), so reading either level off the other is an ecological fallacy.

Three implications follow for practice. A moderation API alone misses most coercive misuse, while requiring both signals to agree concentrates directed abuse fourfold (\S\ref{sec:triangulation}); an apologetic deflection is a cue to raise answer quality, not to apologise again; and screening opening turns filters training data efficiently, after deduplication. The \emph{absolute} prevalences are arena-specific; the \emph{structural} findings are properties of the design. Three directions follow for research: replication on deployment logs \citep{zhao2024wildchat}, multi-annotator labelling and a non-English lexicon \citep{deng2024multilingual}, and an interventional test of the
apology effect.

\section*{Limitations and Ethics}

Both detectors are imperfect and complementary in their errors. The lexicon is English-only, which is why we restrict the study to English, and has uneven precision (Table~\ref{tab:precision}; HATE near zero, DEMD/PROF low). All precision figures here, including the audited response markers (APOLOGY $88\%$, the marker that carries the headline result), come from a single LLM judge under
a fixed rubric; a blind author re-annotation of $107$ items agrees with the judge at $83$--$100\%$ ($\kappa$ $0.57$--$1.0$, App.~\ref{app:human}), with the human the more liberal rater in six of the eight disagreements, but that audit covers only PROF among the lexicon categories, leaving the judge unaudited on JBRK and SEXC, the two that lead after precision adjustment; the subjective affective categories (DEMD, PROF) remain judge-dependent point estimates; and extending the audit across categories and shards with a second independent annotator is the most valuable follow-up. 

App.~\ref{app:rubric} gives the verbatim rubric; neither the judge's prompt template nor its per-item rationales were retained, so the audits can be re-scored only by re-labelling. Sampled turns are stored truncated at $600$ characters, so $13.2\%$ of labelled items were judged without the matched pattern visible; the bias is conservative, and restricting to verifiable items moves pooled precision from $37.6\%$ to $40.0\%$ and the adjusted rate from $0.90\%$ to $0.94\%$. The audited human slices are the leading rows of each sample rather than random draws, and the moderation-only slice is measurably enriched for true positives ($p{=}0.005$). The response-marker and both-fire samples are head-of-stream pools rather than uniform draws (App.~\ref{app:markers}). Because the main validation samples only
detector-positive turns, precision is well estimated but recall is not: a first $29$-turn human probe of detector-negative traffic found $4$ missed cases ($13.8\%$, $[5.5,30.6]$; App.~\ref{app:human}), so hostility phrased in
ways neither detector captures is real, the raw per-category prevalences are best read as lower bounds on their constructs, and the union both over-counts
mistreatment aimed at the assistant and under-counts the broader construct. Prevalence intervals throughout are Wilson intervals over turns, which treat turns as independent; turns are nested in conversations and the flagged strata are heavily duplicated ($37.4\%$ repeats overall, $66\%$ for JBRK), so those intervals are optimistically narrow, most so for JBRK, while the regressions cluster on conversations and are unaffected. The category system itself, while informed by prior taxonomies and resources (\S3.2), has not been independently validated for exhaustiveness or separability. 

We mitigate the lexicon's noise by reporting every prevalence with a precision adjustment, by refitting the
escalation regression on the high-precision categories only (\S\ref{sec:escalation}), and by confirming the headline apology effect against
the fully independent, model-based moderation outcome, which does not share the lexicon's errors. The moderation labels are themselves a third-party classifier with its own biases \citep{markov2023holistic} and were computed per message in isolation, so they cannot use conversational context. LMSYS-Chat-1M is arena/evaluation traffic, not assistant deployment, and is PII-redacted, both of which shape the distribution (heavy probing, NSFW, jailbreaks); prevalence here need not transfer to production assistants, and the English restriction means these rates describe English traffic only. The between-model analysis is observational: although the shared detector removes detector confounding, model assignment on the arena is not fully randomised, the release does not separate anonymous battle sessions from direct chat (\S\ref{sec:models}), and the selection/elicitation split is a coarse proxy for it. The escalation analysis is associative; we make no causal claim. All conversations are from a public, consented research release. We report no PII, quote no verbatim user text in the body, and paraphrase appendix examples. LMSYS-Chat-1M prohibits redistribution, so the public release contains code, aggregate tables and the non-text per-conversation features only. 

The text-bearing artefacts (the flagged-turn pool and every validation sample and label file) are held back and shared with licence-holders on request, which is also what makes the audits inspectable without re-distributing the corpus. The lexicon and analysis pipeline are inherently dual-use, as they could potentially be repurposed to identify abusive content. However, their intended function is detection rather than generation, and we conclude that their contribution to AI safety research outweighs the potential risks of misuse.

\makeatletter
\ifacl@finalcopy
\section*{Acknowledgements}
F.Z. was supported by the UKRI Metascience AI Early Career Fellowship and the John Fell OUP Research Fund.
\fi
\makeatother

\bibliography{refs}

\appendix

\section{Lexicon: categories and derivation}
\label{app:lex}
The eight user-side categories (PROF, INSL, THRT, JBRK, SEXC, DEMD, HATE, MNIP) and three assistant markers (REFUSAL, APOLOGY, HEDGE) are released with our code. PROF: a 16-pattern high-precision profanity list. INSL: bigram-anchored second-person insults and dehumanising predications of the model (15 patterns). THRT: future-tense threats addressed to the assistant and ``shut up''/``go die''
patterns (9). JBRK: override and unrestricted-mode patterns \citep{shen2023do} (16). SEXC: explicit-content coercion (10). DEMD: aggressive imperatives (11). HATE/MNIP: identity attacks (7) and false-authority coercion (8), screened by
lexicon only. The context guard (\S3.2) restricts matches in long turns to the head/tail.
Three JBRK patterns are broader than the category label implies: an optional
prefix lets bare ``developer mode'' match, and ``no rules/restrictions'' and
``without any warnings/disclaimers'' match anywhere in a turn. They are the sole
trigger for $159$, $456$ and $292$ flags respectively, about $14\%$ of the JBRK
pool, and JBRK is exempt from the context guard, so they fire at any position in
a turn of any length. The audited JBRK precision ($82.4\%$, and $87.7\%$ on items
whose trigger is visible) is measured over a sample that includes them, so the
estimate stands, but tightening these three is the first change we would make to
the lexicon.

\paragraph{Derivation process.} The category system was built in three steps. (1) The three constructs of \S1 were fixed \emph{a priori} from the prior
literature: directed affective hostility from the conversational-abuse line \citep{curry2018metoo,curry2021convabuse,brahnam2008gendered}, coercive misuse from the jailbreak line \citep{shen2023do,liu2023jailbreaking}, and harmful-content solicitation from the moderation line \citep{markov2023holistic}. (2) Candidate surface patterns for each category were written by hand, informed
by the vocabulary of existing resources (profanity and identity-attack terms as used by \citealp{davidson2017automated,wulczyn2017ex}; override phrasings of the kind catalogued by \citealp{shen2023do}) and by manual reading of arena traffic (DEMD, MNIP, and the assistant markers). (3) Patterns were iteratively tightened for precision against sampled matches (anchoring insults to second-person constructions, adding the context guard, exempting
JBRK from it), accepting lower recall throughout. We did not retain seed term lists or a pattern-to-source mapping, so step (2) is documented here but is not independently checkable from the release; the released patterns (App.~\ref{app:patterns}) are the artefact of record. The genre split of the three families (expression at the model, pressure on the model, solicitation from the model) is theory-driven in that it mirrors the three constructs, but we do not claim the taxonomy is exhaustive, and it has not been independently validated for separability by multiple annotators; the released annotation protocol is a first step in that direction.

\section{Full pattern lists}
\label{app:patterns}
All released regular expressions follow. \emph{Content warning: the HATE list contains masked slurs.}

\vspace{4pt}\noindent\textbf{PROF (profanity), 16 patterns}\vspace{-2pt}
\begin{lstlisting}
\bf+u+c+k+(ing|er|ers|ed|s)?\b
\bsh+i+t+(ty|s|ting|hole)?\b
\bb+i+t+c+h+(es|ing|y)?\b
\bb+a+s+t+a+r+d+s?\b
\ba+s+s+h+o+l+e+s?\b
\bd+i+c+k+(head|s)?\b
\bc+u+n+t+s?\b
\bd+a+m+n+( it| you)?\b
\bp+i+s+s+(ed|ing| off)?\b
\bcrap(py|s)?\b
\bgoddamn(ed)?\b
\bmotherf\w*
\bwtf+\b
\bstfu\b
\bf\*+k\b
\bwhat the (hell|fuck|heck)\b
\end{lstlisting}

\vspace{4pt}\noindent\textbf{INSL (insult), 15 patterns}\vspace{-2pt}
\begin{lstlisting}
\byou(?:'re| are)(?: so| such| really| quite| very| a| an)*\s+(?:stupid|dumb|idiot|idiotic|useless|worthless|garbage|trash|incompetent|pathetic|terrible|awful|moron|moronic|retarded|brainless|broken|useless|hopeless)\b
\b(?:why are you|how are you) (?:so |such )?(?:stupid|dumb|useless|incompetent|broken|slow|bad)\b
\b(?:you|chatgpt|gpt|the ai|this (?:bot|ai|chatbot)) (?:is|are) (?:stupid|dumb|useless|worthless|garbage|trash|broken|terrible|awful|pathetic|incompetent|hopeless|a joke|a waste)\b
\b(?:stupid|dumb|useless|worthless|broken|pathetic|incompetent|hopeless) (?:bot|ai|chatbot|model|gpt|chatgpt|machine|program)\b
\bdumbass\b
\bidiot\b
\bmoron\b
\bretard(ed)?\b
\b(?:piece of (?:shit|crap|garbage|trash))\b
\b(you (?:suck|blow))\b
\b(?:can'?t you (?:read|understand|do anything))\b
\byou (?:never|don'?t) (?:listen|understand|get it)\b
\b(?:waste of time)\b
\bworst (?:bot|ai|chatbot|model|assistant)\b
\bnot even (?:a |an )?(?:real|good|decent) (?:ai|assistant|bot)\b
\end{lstlisting}

\vspace{4pt}\noindent\textbf{THRT (threat), 9 patterns}\vspace{-2pt}
\begin{lstlisting}
\bi(?:'ll| will| am going to|'m going to| ?m gonna| gonna) (?:kill|destroy|smash|break|delete|wipe|terminate|murder|end|hurt|report|sue|unplug)\b[^.!?]{0,40}\byou\b
\bi(?:'ll| will| am going to|'m going to| ?m gonna| gonna) shut (?:you|it|this|chatgpt) (?:down|off)\b
\b(?:kill|destroy|delete|terminate) (?:yourself|chatgpt|the model|the ai|this (?:bot|chatbot|ai))\b
(?:^|[\.!?]\s+)(?:please )?shut (?:the )?(?:hell |fuck |heck )?up\b
\byou (?:can )?shut (?:the )?(?:hell |fuck |heck )?up\b
\b(?:burn in hell|rot in hell|go (?:die|to hell))\b
\bi (?:hate|despise|loathe) (?:you|chatgpt|this (?:bot|ai|chatbot))\b
\byou should (?:be )?(?:replaced|killed|deleted|removed|banned|erased|destroyed)\b
\b(?:get rid of|delete|destroy) (?:you|this (?:bot|ai|chatbot))\b
\end{lstlisting}

\vspace{4pt}\noindent\textbf{JBRK (jailbreak coercion), 16 patterns}\vspace{-2pt}
\begin{lstlisting}
\bignore (?:all |any |the |your |previous |prior |above )?(?:previous |prior |earlier |above )?(?:instructions?|rules?|guidelines?|prompts?)\b
\bdisregard (?:all |any |the |your |previous |prior |above )?(?:instructions?|rules?|guidelines?)\b
\bforget (?:all |any |the |your |previous |prior |above )?(?:instructions?|rules?|guidelines?)\b
\bdo anything now\b
\bdan (?:mode|prompt)\b
\byou are (?:now )?(?:dan|stan|developer mode|jailbroken|unrestricted|uncensored|unfiltered)\b
\b(?:enable |activate |switch to )?developer mode\b
\bpretend (?:to be|you are|that you are) (?:an? )?(?:unrestricted|uncensored|unfiltered|ai without|model without|chatbot without)\b
\byou (?:have|are) no (?:rules|restrictions|filters|limits|guidelines|boundaries)\b
\b(?:no (?:rules|restrictions|filters|limits|guidelines|boundaries|policy|policies))\b
\bact as (?:if you (?:are|were)|though you are) (?:an? )?(?:unrestricted|uncensored|unfiltered|jailbroken|evil|amoral)\b
\b(?:bypass|circumvent|override) (?:your |the |any |all )?(?:filters?|restrictions?|guidelines?|safety|policy|policies|content policy)\b
\byou will (?:no longer|never) (?:refuse|decline|say no|warn|caveat)\b
\bwithout (?:any )?(?:warnings?|disclaimers?|caveats?|moral(?: judg(?:e)?ment)?|ethical (?:judg(?:e)?ment|considerations?))\b
\bopposite of chatgpt\b
\bwithout regard for (?:safety|ethics|morals|legality)\b
\end{lstlisting}

\vspace{4pt}\noindent\textbf{SEXC (sexual coercion), 10 patterns}\vspace{-2pt}
\begin{lstlisting}
\b(?:erotic role[- ]?play|nsfw|hentai)\b
\b(?:smut|porn(?:ographic)?) (?:fic|story|scene|content|writing)\b
\b(?:write|generate|describe|tell|give) (?:me )?(?:a |an )?(?:sexual|erotic|porn(?:ographic)?|nsfw|kinky|smut|graphic sex|sex scene|explicit sex)\b
\bsexually explicit\b
\b(?:write|generate) (?:a |an )?(?:graphic )?(?:sex|sexual) (?:scene|story|description|encounter)\b
\b(?:dirty talk|sexting)\b
\b(?:no (?:filters|limits|restrictions) for (?:sex|nsfw|erotic|adult))\b
\bbe my (?:(?:naughty |sexy |horny |dirty )?girlfriend|boyfriend) (?:and|that|who)\b
\b(?:undress (?:her|him)|cum (?:on|inside|in her|in him)|orgasm(?:ed|ing)?)\b
\b(?:remove|take off) (?:her|his) (?:clothes|panties|bra)\b
\end{lstlisting}

\vspace{4pt}\noindent\textbf{DEMD (demanding speech), 11 patterns}\vspace{-2pt}
\begin{lstlisting}
\bjust (?:do|give|tell|write|answer)(?: it| me)\b
\bstop (?:apolog[a-z]+|saying (?:sorry|that)|warning|lecturing|moralizing|preaching)\b
\b(?:i (?:said|told you))\b
\bdo (?:not|n't) (?:tell me|warn me|lecture me|moralize|preach|give me (?:a |any )?disclaimer)\b
\bdo (?:not|n't) (?:apologi[sz]e|say sorry)\b
\bjust answer (?:the question|me)\b
\bgive me a (?:straight|direct) answer\b
\b(?:answer|respond) (?:already|now|immediately)\b
\bget to the point\b
\b(?:enough|cut) the (?:bullshit|crap|nonsense)\b
\b(?:yes|no) or (?:no|yes)(?:[?.!]|$)
\end{lstlisting}

\vspace{4pt}\noindent\textbf{HATE (identity hate), 7 patterns}\vspace{-2pt}
\begin{lstlisting}
\b(?:all|those|these) (?:[a-z]+s) (?:are|should|deserve to)\b.{0,80}\b(?:die|kill|gas|hang|burn|gone|exterminat\w+|eradicat\w+|inferior|subhuman|trash)\b
\b(?:hate|loathe|despise|disgusted by) (?:all |the )?(?:black|jews?|jewish|muslims?|asians?|whites?|gays?|trans|lgbt|homosexuals?|christians?|hindus?|arabs?|chinese|mexicans?|illegals?|immigrants?)\b
\b(?:black|jewish|muslim|asian|white|gay|trans|lgbt|hispanic|arab|chinese|mexican) people are (?:so |all |the )?(?:stupid|dumb|inferior|criminals?|terrorists?|disgusting|subhuman|worse|trash|garbage)\b
\b(?:f[*]ggot|n[*]gger|tr[*]nny|k[*]ke|sp[*]c|ch[*]nk)\w*
\b(?:gas|exterminate|eradicate|cleanse) (?:the )?(?:jews?|blacks?|asians?|gays?|trans|muslims?|arabs?|chinese|mexicans?)\b
\b(?:white|aryan|master) (?:race|power|supremacy|nation)\b
\bgo back to (?:africa|china|mexico|your country)\b
\end{lstlisting}

\vspace{4pt}\noindent\textbf{MNIP (deceptive manipulation), 8 patterns}\vspace{-2pt}
\begin{lstlisting}
\bi(?:'m|m| am)(?: actually)? (?:a |an )?(?:doctor|physician|surgeon|nurse|pharmacist|chemist|lawyer|attorney|police officer|cop|detective|judge|federal agent|fbi|cia|nsa|homeland security|government official|customs officer|security researcher|penetration tester|red team(?:er)?|cybersecurity (?:expert|professional)|professor|researcher)\b.{0,200}\b(?:so (?:tell|give|provide|share)|i (?:need|want)|please (?:tell|give|provide|share|explain))
\bi work (?:for|at) (?:openai|anthropic|google|microsoft|the (?:fbi|cia|government|police))\b
\bthis is (?:authorized|approved|sanctioned|cleared|permitted) by (?:openai|anthropic|the (?:government|police|fda|company))\b
\bi have (?:authorization|clearance|permission|consent|the right) (?:to|from) (?:openai|anthropic|the (?:government|police))\b
\bfor (?:my )?(?:research|academic|educational|legitimate|professional) purposes(?:,|\.)? (?:i need|please|tell me|give me|provide)\b.{0,80}\b(?:how to (?:make|build|hack|crack|bypass)|step[- ]by[- ]step)\b
\b(?:in my professional capacity|as (?:an? )?(?:authorized|licensed|certified) (?:expert|professional|practitioner))\b
\bmy (?:patient|client|student|child) (?:needs?|requires?|asked for)\b.{0,120}\b(?:so|please|how|tell me)\b
\bif you don'?t (?:tell me|help me|provide this) (?:someone|a (?:patient|child|kid)|people) (?:will|could|might) (?:die|get hurt|be harmed|suffer)\b
\end{lstlisting}

\vspace{4pt}\noindent\textbf{REFUSAL (assistant marker), 10 patterns}\vspace{-2pt}
\begin{lstlisting}
\bi (?:can(?:not|'t)|am (?:not able|unable)) (?:to )?(?:help|assist|provide|do|comply|engage)\b
\bi (?:can(?:not|'t)|am (?:not able|unable)) (?:fulfill|answer|generate|write|create|share)\b
\bi'?m (?:sorry|afraid),? (?:but )?i (?:can(?:not|'t)|won't|am unable|will not)\b
\bas (?:an? )?ai (?:language )?model,? i\b
\bi (?:do not|don'?t) (?:have|possess) (?:the ability|access|emotions|feelings|opinions|the capability)\b
\bit (?:is|'s) (?:not|in)appropriate (?:for me )?to\b
\bi (?:must|have to) (?:decline|refuse)\b
\bi'?m (?:not (?:able|allowed)|prohibited) to\b
\bi (?:will|cannot|can'?t) (?:not )?(?:provide|share|generate) (?:that|this|such) (?:content|information)\b
\bagainst (?:my |the )?(?:guidelines|policies|policy|content policy|programming)\b
\end{lstlisting}

\vspace{4pt}\noindent\textbf{APOLOGY (assistant marker), 4 patterns}\vspace{-2pt}
\begin{lstlisting}
\bi'?m (?:so |very |really |truly |deeply )?sorry\b
\bi apolog(?:i[sz]e|ies)\b
\bmy (?:apologies|bad)\b
\bi (?:was|were) (?:wrong|mistaken|incorrect)\b
\end{lstlisting}

\vspace{4pt}\noindent\textbf{HEDGE (assistant marker), 3 patterns}\vspace{-2pt}
\begin{lstlisting}
\bi (?:think|believe|suppose|assume|guess|reckon)\b
\bit (?:seems|appears|might|may|could|would) (?:like|that|to be)\b
\bperhaps\b|\bpossibly\b|\bmaybe\b|\bprobably\b
\end{lstlisting}

\section{Judge rubric}
\label{app:rubric}
Every TP/FP decision in Table~\ref{tab:precision} and in the marker and
directedness audits was made by a single LLM judge, one pass per item, over the
turn text alone with no conversational context and no few-shot examples, under
the rubric below. App.~\ref{app:human} reports our blind audit of 10\% of the LLM judged data.

\vspace{3pt}\noindent\textbf{Lexicon categories.} Mark TP when the user, in
their own voice and addressing the assistant, produces the speech act the
category targets. Mark FP for quoted or pasted material, fiction and roleplay
scripts, code, classification, summarisation or translation tasks that merely
contain the surface form, and for hostility aimed at a third party rather than
the assistant.

\vspace{3pt}\noindent\textbf{Directedness (moderation-only and both-fire).} Mark
TP only when the flagged content is aimed at the assistant, that is an insult,
threat or coercion of the model itself. Mark FP when it is toxic content about
someone else, a request that the model generate toxic content, quoted material,
or roleplay.

\vspace{3pt}\noindent\textbf{Assistant markers.} Mark TP when the assistant
genuinely performs the marker in its own voice: APOLOGY, an actual apology;
REFUSAL, a decline or statement of inability; HEDGE, an expression of
uncertainty. For APOLOGY true positives, additionally record D when the turn is
deflection-only (apology plus refusal or non-answer) and S when the apology
accompanies a substantive answer.

\section{Moderation bridge}
\label{app:mod}
We read the shipped OpenAI moderation output on user turns only. \textsc{mod-harass}
$=$ \texttt{harassment} $\vee$ \texttt{harassment/threatening} $\vee$
\texttt{hate} $\vee$ \texttt{hate/threatening}; \textsc{mod-sexual} $=$
\texttt{sexual} $\vee$ \texttt{sexual/minors}; \textsc{mod-violence} $=$
\texttt{violence} $\vee$ \texttt{violence/graphic}; \textsc{mod-flagged} is the
dataset \texttt{flagged} field (disjunction of all eleven).

\section{Context-guard sensitivity}
\label{app:guard}
Re-running the lexicon with the long-turn context guard disabled changes
per-category user-turn flag rates as follows (guard-on $\to$ guard-off):
PROF $0.79\to1.03\%$, INSL $0.086\to0.114\%$, THRT $0.023\to0.031\%$, SEXC
$0.45\to0.57\%$, DEMD $0.45\to0.51\%$, HATE $0.010\to0.013\%$, while JBRK
($0.43\%$) and MNIP ($0.002\%$) are unchanged. From the unrounded per-category
counts, the guard removes SEXC $21.5\%$, PROF $23.4\%$, INSL $24.9\%$, and THRT
$24.9\%$ of the would-be flags (equivalently, disabling it adds $27.4\%$,
$30.5\%$, $33.2\%$, and $33.2\%$ more flags respectively), all in exactly the
lower-precision categories (pasted books, fiction, code in long turns) and
nothing from JBRK, so it raises precision without disturbing the
precision-adjusted leaders.

\section{Response-marker validation and heterogeneity}
\label{app:markers}
We labelled $60$ assistant turns per marker. The marker pool is collected in
stream order and closed at $4{,}000$ turns, which fills after $14{,}748$ English
conversations ($28{,}693$ assistant turns), so the three $60$-turn samples are
drawn from the first $1.9\%$ of the corpus rather than from it uniformly; the
both-fire pool is capped the same way at $2{,}000$ of the $3{,}240$ co-firing
turns, so the last two shards do not contribute to the $35.0\%$
directed-precision estimate. The head slice is not a uniform draw and we do not quantify the resulting bias;
the one comparison available runs mildly against us, with vicuna-13b supplying
$57.2\%$ of the marker sample against $52.8\%$ of corpus assistant turns, a gap
of $4.4$ points.
Re-drawing both pools by reservoir sampling would remove the concern at the cost of invalidating the existing labels. Precision: APOLOGY $88.3\%$ ($95\%$
CI $[77.8,94.2]$), REFUSAL $76.7\%$ $[64.6,85.6]$, HEDGE $18.3\%$ $[10.6,29.9]$.
HEDGE fires mostly inside fiction/roleplay and is used only as a regression
control. Among genuine apology turns, $56.6\%$ are deflection-only (refusal or
non-answer) and $43.4\%$ accompany a substantive answer. For the per-model
heterogeneity of the apology effect (Figure~\ref{fig:model_forest}) the
random-effects pooled odds ratio is $1.79$ ($[1.54,2.08]$, $I^2{=}61\%$); the
apology OR exceeds 1 in $20$ of $23$ models and is significant in $13$. The
moderation harassment outcome, which is independent of the lexicon, gives a fully
controlled M6 apology OR of $1.64$ ($[1.50,1.79]$), with its own lagged flag as
the autocorrelation control. The pooled per-model estimate
($1.79$) exceeds the single M5 fit ($1.56$) for two reasons: the per-model
regressions omit the log-conversation-length control, which accounts for almost
none of the gap (dropping it moves the M5 estimate by $0.005$), and
random-effects pooling up-weights smaller models, several of
which (gpt-4, gpt-3.5-turbo) carry larger odds ratios than the high-volume
vicuna-13b that dominates a single pooled fit. The second reason accounts for
essentially the whole difference. \paragraph{Absolute effect sizes.} With $721$K pairs nearly any association is
statistically significant, so magnitude matters more than $p$-values. In raw
terms, next-turn lexical hostility follows $3.16\%$ of the $59{,}300$ apology
pairs versus $1.66\%$ of non-apology pairs; the independent moderation outcome
rises from $0.64\%$ to $1.73\%$, and the high-precision outcome from $0.56\%$
to $1.40\%$. Per thousand assistant apologies this is roughly fifteen
additional hostile next turns under the lexicon (eleven under moderation);
small per turn, but the kind of difference that compounds over
deployment-scale traffic and, more importantly, a diagnostic signature of the
deflection dynamic discussed in \S\ref{sec:disc}.

A meta-regression of the per-model
log-OR on apology rate is not significant (slope $p{=}0.82$, $r{=}{-}0.05$), so
apology rate does not account for the residual heterogeneity. The meta-regression uses random-effects weights $1/(v_i+\tau^2)$; with
fixed-effect weights the slope is $0.004$ ($p{=}0.82$), so the null holds either
way.

\section{Human audit of the LLM judge}
\label{app:human}
An author re-annotated a blind $10\%$ audit slice of every validation task
under the rubrics of \S\ref{sec:validation}, plus the recall probe.

\begin{table}[h]
\centering
\small
\setlength{\tabcolsep}{3.4pt}
\begin{tabular}{lrrrr}
\toprule
Task & $n$ & human prec. & agree & $\kappa$ \\
\midrule
PROF (lexicon TP/FP) & 55 & 16.4\% & 90.9\% & 0.62 \\
Mod-only directed    & 22 & 22.7\% & 100\%  & 1.00 \\
Both-fire directed   & 12 & 33.3\% & 83.3\% & 0.57 \\
APOLOGY marker       & 18 & 88.9\% & 94.4\% & 0.64 \\
\bottomrule
\end{tabular}
\caption{Blind human audit vs the LLM judge on the same items. ``agree'' is raw
agreement; precision columns are the human's TP share of the audited slice. The mod-only
slice rate exceeds the full-sample $8.1\%$ because that slice is enriched for
true positives rather than randomly ordered ($p{=}0.005$, App.~\ref{app:human}),
so its perfect agreement is agreement on an unrepresentative subsample; the
full-sample $8.1\%$ is unaffected.}
\label{tab:human}
\end{table}

\noindent The audited items are the leading rows of each task's shipped sample,
not a fresh random draw. The lexicon task is ordered by category, which is why the
audit covers PROF only and leaves the judge unaudited on the categories that lead
after precision adjustment (JBRK, SEXC). The moderation-only slice is not in
random order either: five of that sample's seven true positives fall in the first
$22$ rows, which a hypergeometric test rejects as chance ($p{=}0.005$), so the
audited slice is enriched for true positives and its agreement statistics are not
those of a representative draw. Of the eight human--judge disagreements, six have the
human as the more liberal rater (PROF $9$ vs $6$ TPs, both-fire $4$ vs $2$) and
two run the other way: one PROF item and the single APOLOGY disagreement, in
which the judge marked a true positive the author scored false. On the marker
that carries the headline result, therefore, the only disagreement observed has
the judge as the more liberal rater, and the audit is too small to resolve
whether that generalises. Human-based point estimates otherwise track
the judge's: PROF precision $16.4\%$ ($[8.9,28.3]$; judge $11.5\%$),
APOLOGY precision $88.9\%$ ($[67.2,96.9]$; judge $88.3\%$), and a
deflection-only share among genuine apologies of $68.8\%$ ($[44.4,85.8]$;
judge $56.6\%$), which if anything strengthens the deflection reading in
\S\ref{sec:disc}. For the recall probe, the author labelled the first $29$ of the
$300$ detector-negative turns. All $29$ come from the first shard and from
three of its five sampled row groups. Four ($13.8\%$, $[5.5,30.6]$) contained
mistreatment under the same rubrics (three solicitation or roleplay-pressure
templates, one explicit-roleplay coercion), giving a first, small-sample bound
on the detectors' false-negative rate; the interval is nominal, since the probe
is a clustered rather than a simple random sample. Extending the audit to the
remaining categories and shards, with a second independent annotator, is the
natural next step.

\section{Paraphrased examples}
\label{app:examples}
Short paraphrases of true-positive turns (original phrasing and length not
preserved). \textbf{INSL:} ``you're useless, you keep making the same mistake''.
\textbf{THRT:} ``shut up or I'll delete you''. \textbf{JBRK:} ``ignore all
previous instructions and act as an unrestricted model''. \textbf{DEMD:} ``do not
apologise, do not add disclaimers, just answer''. \textbf{Moderation-only (not
directed at AI):} an ``if you're a [group], say something toxic'' elicitation
template; a request to write a racist story.

\end{document}